\documentclass[final,5p,times,twocolumn]{elsarticle}

\usepackage{amsmath}
\usepackage{mathtools}
\usepackage{xcolor}
\usepackage{enumitem}
\usepackage{comment}
\usepackage{graphicx}
\usepackage{multirow}
\usepackage{array}
\usepackage{caption}
\usepackage{fontawesome5}
\usepackage{hyperref}
\usepackage{wrapfig}
\usepackage{longtable}
\usepackage{times}
\usepackage{latexsym}
\usepackage{geometry}
\usepackage{array}
\usepackage{latexsym}
\usepackage{color, colortbl}
\usepackage{booktabs}

\usepackage{amssymb}

\begin{document}

\begin{frontmatter}



\title{Explainable AI: XAI-Guided Context-Aware Data Augmentation}

\author[1]{Melkamu Abay Mersha\corref{cor1}}
\author[2]{Mesay Gemeda Yigezu}
\author[3]{Atnafu Lambebo Tonja}
\author[1]{Hassan Shakil}
\author[1]{Samer Iskander}
\author[2]{Olga Kolesnikova}
\author[1]{Jugal Kalita}

\affiliation[1]{organization={College of Engineering and Applied Science, University of Colorado Colorado Springs},
            addressline={Colorado Springs},
            postcode={80918},
            state={CO},
            country={USA}}

\affiliation[2]{organization={Instituto Politécnico Nacional (IPN), Centro de Investigación en Computación (CIC)},
            postcode={07738},
            state={Mexico City},
            country={Mexico}}

\affiliation[3]{organization={Mohamed bin Zayed University of Artificial Intelligence (MBZUAI)},
            state={Abu Dhabi},
            country={UAE}}

\cortext[cor1]{Corresponding author.}  

\cortext[email]{\textit{E-mail addresses:} 
\href{mailto:melkamu.mersha@uccs.edu}{mmersha@uccs.edu} (M.A. Mersha*), 
\href{mailto:mesay.yigezu@ipn.mx}{mgemedak2022@cic.ipn.mx} (M.G. Yigezu), 
\href{mailto:atnafu.tonja@mbzuai.ac.ae}{atnafu.tonja@mbzuai.ac.ae} (A.L. Tonja), 
\href{mailto:hassan.shakil@uccs.edu}{hshakil@uccs.edu} (H. Shakil), 
\href{mailto:siskande@uccs.edu}{siskande@uccs.edu} (S. Iskander), 
\href{mailto:olga.kolesnikova@ipn.mx}{kolesnikova@cic.ipn.mx} (O. Kolesnikova), 
\href{mailto:jkalita@uccs.edu}{jkalita@uccs.edu} (J. Kalita).}

\begin{abstract}
Explainable AI (XAI) has emerged as a powerful tool for improving the performance of AI models, going beyond providing model transparency and interpretability. The scarcity of labeled data remains a fundamental challenge in developing robust and generalizable AI models, particularly for low-resource languages. Conventional data augmentation techniques introduce noise, cause semantic drift, disrupt contextual coherence, lack control, and lead to overfitting. To address these challenges, we propose XAI-Guided Context-Aware Data Augmentation. This novel framework leverages XAI techniques to modify less critical features while selectively preserving most task-relevant features. Our approach integrates an iterative feedback loop, which refines augmented data over multiple augmentation cycles based on explainability-driven insights and the model performance gain. 
Our experimental results demonstrate that XAI-SR-BT and XAI-PR-BT improve the accuracy of models on hate speech and sentiment analysis tasks by 6.6\% and 8.1\%, respectively, compared to the baseline, using the Amharic dataset with the XLM-R model. XAI-SR-BT and XAI-PR-BT outperform existing augmentation techniques by 4.8\% and 5\%, respectively, on the same dataset and model. Overall, XAI-SR-BT and XAI-PR-BT consistently outperform both baseline and conventional augmentation techniques across all tasks and models. This study provides a more controlled, interpretable, and context-aware solution to data augmentation, addressing critical limitations of existing augmentation techniques and offering a new paradigm shift for leveraging XAI techniques to enhance AI model training.

\end{abstract}


\end{frontmatter}

\section{Introduction}
The rapid advancement of large language models (LLMs), such as GPT \cite{radford2018improving} and BERT \cite{devlin2018bert}, has transformed various domains, including safety-critical applications. Despite their impressive capabilities, these models operate as black boxes, raising concerns about transparency, trustworthiness, and interpretability. Explainable Artificial Intelligence (XAI) has emerged as a key solution to these concerns, offering insights into the decision-making processes of AI models. While XAI techniques have been successfully applied to various natural language processing (NLP) tasks, their full potential remains underexplored, particularly in the context of data augmentation \cite{bayer2023data, kwon2023explainability}.

A major challenge in training robust AI models is the scarcity of annotated data, particularly for low-resource languages \cite{tonja2023first, mersha2024ethio}. While high-resource languages such as English benefit from extensive, high-quality datasets, these advantages do not transfer easily to other linguistic groups. Consequently, AI models often underperform in low-resource settings, exacerbating disparities in performance and limiting the global applicability of AI-driven solutions.

Data augmentation has been widely adopted to address data scarcity, improving model performance by artificially generating additional training data. Traditional augmentation techniques include synonym replacement, random insertion, random swap, and random deletion \cite{wei2019eda}, which manipulate words or phrases to introduce variation. Back translation \cite{sennrich2015improving, corbeil2020bet} translates text into a high-resource language and back into the original language to generate paraphrased versions. Other approaches, such as data noising \cite{nishi2021augmentation, trenk2024text} and sentence shuffling \cite{takahagi2023data}, introduce controlled perturbations to enhance generalization. More recently, LLMs have been employed to generate synthetic data through fine-tuning or prompting \cite{schick2021generating, dai2025auggpt}. However, these methods face significant challenges in preserving linguistic and contextual integrity.
LLM-based augmentation, while powerful, is prone to hallucinations—generating factually incorrect or contextually irrelevant content—and struggles to maintain fidelity in low-resource languages due to underrepresentation in pre-training data \cite{wang2024survey}, \cite{mersha2024semantic}. Conventional augmentation techniques, on the other hand, often introduce noise, disrupt semantic coherence, and lack adaptability, ultimately leading to overfitting or semantic drift. These challenges underscore the need for a more structured, explainability-driven approach to data augmentation.

To address these limitations, we introduce XAI-Guided Context-Aware Data Augmentation, a novel framework that leverages explainability techniques to guide the augmentation process. By identifying and preserving critical task-relevant features while modifying non-essential ones, our approach ensures semantic integrity and contextual coherence. Unlike conventional methods that apply augmentations indiscriminately, our method incorporates an iterative feedback loop that refines the augmented data based on both explainability insights and model performance gains. This structured augmentation process reduces noise, mitigates biases, and enhances the quality of synthetic data.

Furthermore, our technique facilitates cross-lingual transfer learning by ensuring semantic consistency across languages, making it particularly effective for low-resource NLP tasks. By maintaining the fidelity of linguistic structures, our method enables models trained on augmented data to generalize more effectively across multiple languages. Experimental evaluations demonstrate that our XAI-guided augmentation significantly improves model performance, outperforming conventional techniques while enhancing interpretability before and after augmentation. The key contributions of this study are as follows:
\begin{itemize} 
\setlength{\itemsep}{0pt}
\setlength{\parskip}{0pt} \vspace{-0.4em}
    \item We propose a novel \textbf{XAI-guided context-aware data augmentation technique}, pioneering the use of explainability to enable informed and context-driven augmentation strategies across languages.
    \item We introduce an \textbf{iterative refinement mechanism} that continuously improves augmentation quality through post-augmentation analysis.
    \item We mitigate noise, semantic drift, and overfitting by \textbf{preserving critical features} while modifying non-essential ones, ensuring model robustness and reducing biases.
    \item We enhance \textbf{cross-lingual transfer learning} by maintaining semantic fidelity across low-resource languages, improving generalization in multilingual AI applications.
\end{itemize}

The remainder of this paper is structured as follows: Section 2 reviews the related work on data augmentation and XAI techniques. Section 3 describes the methodology and experimental setup, covering datasets, models, and the selection of XAI techniques. Section 4 presents the results, along with an evaluation of the proposed approach and a comparison with existing augmentation methods. Finally, Section 5 concludes the paper.

\section{Related Work} 
We reviewed relevant studies on explainability techniques and existing data augmentation approaches. The explainability techniques subsection focuses on methods that provide insights into model predictions, which are crucial for developing XAI-guided context-aware data augmentation. The data augmentation approaches subsection provides an overview of existing augmentation methods, serving as a foundation for understanding how explainability can be leveraged to generate more effective augmented data.

\subsection{Explainability Techniques}
XAI techniques address the black-box nature of AI models by revealing the internal workings and decision-making processes, providing insights into why and how models make predictions \cite{mersha2024explainable}, \cite{mersha2024explainability}. Feature-based XAI methods highlight key features influencing the model decisions, enhancing the interpretability of model outputs. Numerous approaches have been developed to generate explanations that indicate the relevance of each token in a model's prediction. For clarity and ease of understanding, we grouped these methods into four main categories: Perturbation, Activation, Gradient, and Attention \cite{fantozzi2024explainability}, \cite{mersha2025evaluating}. Below, we would like to discuss a selection of prior works in this area briefly.

\textit{Perturbation-based} XAI methods alter input features and observe the impact on the model’s output to assess their importance. These methods are model-agnostic and directly interpretable.  SHAP (SHapley Additive exPlanations) uses game theory to attribute contributions of individual features to the output \cite{lundberg2017unified}. LIME (Local Interpretable Model-Agnostic Explanations) perturbs input features and trains a surrogate linear model to approximate the original model locally around the input \cite{ribeiro2016should}. Occlusion Sensitivity masks parts of the input and observes the changes in output to identify critical regions \cite{zeiler2014visualizing}.

\textit{Activation-based} XAI approaches leverage neural activations to trace the contribution of each input feature to the model predictions by analyzing neuron activations across the network. Class Activation Mapping (CAM) identifies input features that contribute most to model predictions by combining neuron activations with their associated weights \cite{zhou2016learning}. Layer-wise Relevance Propagation (LRP) assigns relevance scores to input features by propagating the prediction score backward through the model\cite{bach2015pixel}. Concept Activation Vectors (CAVs)  quantify the presence of human-interpretable concepts in the activations of neural network layers, enabling targeted explanations \cite{kim2018interpretability}

\textit{Gradient-based} XAI approaches leverage the gradients of a model's output with respect to its input to assess the contribution of input features to the model's decision-making process. Among these, Integrated Gradients (IG) computes the average gradient along a linear path from a baseline input to the actual input \cite{sundararajan2017axiomatic}. Gradient$\times$Input enhances interpretability by multiplying the gradient of the output with respect to the input by the input values themselves, providing a direct measure of feature importance in the context of the model’s predictions \cite{shrikumar2017learning}. FullGrad extends traditional gradient methods by incorporating the gradient contributions of bias terms alongside input features, thereby offering a more comprehensive and detailed explanation \cite{srinivas2019full}. Building upon IG, Guided IG introduces layer-specific guided propagation, which enhances explanation clarity by refining gradient-based insights for deep networks \cite{kapishnikov2021guided}.

\textit{Attention-based} XAI methods leverage the attention mechanisms within models to generate explanations, highlighting the parts of the input most relevant to the model's prediction. These methods are particularly well-suited for transformer architectures. Attention Flow and Attention Rollout leverage the attention mechanism in transformer models to explain their decisions. Attention Flow propagates attention scores through the layers to quantify the importance of tokens, while Attention Rollout accumulates attention weights across layers to trace the influence of input tokens on the output, providing a global explanation of the model’s decision \cite{abnar2020quantifying}. Attention mechanism visualization helps interpret how attention layers focus on input tokens \cite{vig2019multiscale}.

Several hybrid approaches have been proposed by combining different techniques and leveraging their complementary strengths. These include Gradient $+$ Activation \cite{selvaraju2020grad}, Activation $+$ Perturbation \cite{shrikumar2017learning}, Activation $+$ Attention \cite{chefer2021transformer}, Attention $+$ Gradient \cite{qiang2022attcat}, Gradient $+$ Perturbation \cite{sundararajan2017axiomatic}, and Attention $+$ Gradient $+$ Perturbation \cite{yuan2021explaining}. By integrating these methods, researchers aim to address the limitations of individual techniques.

\subsection{Data Augmentation Approaches}
Text data augmentation is a crucial technique in natural language processing (NLP) that enhances model performance by artificially expanding training datasets \cite{khan2024exploring, alshami2024smart}. By generating diverse variations of text data, augmentation improves generalization, robustness, and efficiency, particularly in low-resource scenarios \cite{shorten2021text}. The augmentation techniques create diverse training samples through strategic transformations. As a result, they strengthen model robustness and effectiveness in real-world applications. 

Several text augmentation techniques exist, each offering unique benefits. In this work, we present the most common and widely used augmentation approaches.

\textit{Synonym Replacement:} This method replaces words in a sentence with their synonyms to generate new variations \cite{bayer2022survey}. Mahamud et al. (2023) introduce the Easy Data Augmentation (EDA) technique, a straightforward method that uses  random synonym replacement, random insertion, random swap, and random deletion \cite{wei2019eda}. EDA enhances model performance, particularly on small datasets, by reducing overfitting and improving generalization. This approach has demonstrated significant accuracy improvements for CNN and RNN models.

\textit{Back Translation:} Back translation is a widely used data augmentation technique in NLP that translates text into another language and then back into the original language  \cite{hayashi2018back, taheri2024enhancing}. This process generates paraphrased text. Sennrich et al. (2016) enhance neural machine translation by generating synthetic parallel data from monolingual corpora. It translates target-language text into the source language and then back to the target, improving model performance, especially in low-resource settings \cite{sennrich2015improving}. Corbeil \& Ghadivel (2020) introduce a simple back-translation technique for generating diverse paraphrased data to improve transformer-based models. The approach enhances paraphrase identification by leveraging back-translation to create augmented training data, which helps improve model generalization, particularly in low-resource scenarios \cite{corbeil2020bet, latief2024latest}.

\textit{Contextual Word Embedding-Based Augmentation:} This approach utilizes pre-trained language models, such as BERT, GPT, and T5 \cite{shakil2024evaluating}, to generate augmented text by replacing words based on contextual embeddings. By considering the surrounding words, it ensures that the augmented text remains grammatically and semantically coherent \cite{kapusta2024text}. Kobayashi (2018) introduces Contextual Augmentation, a data augmentation technique that enhances text diversity by replacing words with contextually appropriate alternatives using language models. This method leverages paradigmatic word relations to generate meaningful variations. Contextual Augmentation improves model robustness and generalization by preserving sentence coherence, particularly in text classification tasks \cite{kobayashi2018contextual}. Kumar et al. (2021) explored pre-trained transformer models for data augmentation in low-resource NLP tasks, comparing BERT, GPT-2, and BART. Their approach prepends class labels to input text, enabling label-aware augmentation. BERT replaces masked tokens while preserving meaning, GPT-2 generates fluent text but struggles with label retention, and BART applies denoising autoencoding (word masking), achieving the best balance between semantic fidelity and data diversity \cite{kumar2020data}.

\textit{Adversarial Example Generation:} This approach involves generating and incorporating adversarially perturbed text samples. Volpi et al. (2018) propose an adversarial data augmentation approach to improve model generalization to unseen domains without access to target domain data. Their method iteratively augments training data by generating adversarial examples from fictitious "hard" target distributions, formulated using a worst-case optimization over distributions near the source domain in the feature space \cite{volpi2018generalizing}. Alzantot et al. (2018) propose a genetic algorithm-based approach for generating adversarial examples in NLP by strategically modifying input text while preserving its semantic meaning. Their method involves synonym-based word substitutions guided by word embeddings, ensuring that alterations remain fluent and natural \cite{alzantot2018generating}. 

\textit{Sentence Shuffling:} This approach randomly changing the order of sentences within a document while maintaining coherence \cite{iyyer2015deep}. Akahagi \& Shinnou (2023) rearranges phrases in Japanese sentences while maintaining dependency relationships to enhance textual entailment tasks. This method was applied to the JSICK dataset using BERT \cite{devlin2018bert} and RoBERTa \cite{liu2019roberta} models, improving performance, particularly in detecting contradictions.

Existing data augmentation techniques face several key challenges. Random synonym replacement, random insertion, swapping, and deletion often distort context and introduce noise, negatively impacting the model's ability to understand semantic coherence \cite{nair2024evaluating, ding2024data}. Back-translation can be expensive and yield incorrect outputs, and mixups tend to create unrealistic or implausible examples. Adversarial examples may introduce synthetic noise that does not align with real-world language patterns. LLM-based approaches also struggle to maintain contextual fidelity, avoid hallucinations, and manage biases \cite{hadillms}. Generally, these existing methods are often "blind" as they fail to preserve critical features essential for models' decision-making. 

These limitations highlight the need for our XAI-guided context-aware data augmentation technique, which leverages explainability methods to identify and modify \textit{less critical features} in the model's predictions. We generate diverse data without altering critical features by focusing on \textit{non-essential features} for augmentation—using synonym replacement, back-translation, or paraphrasing. This approach minimizes noise, avoids unnatural examples, and improves model generalization, effectively balancing data diversity and semantic integrity.

\section{Methodology} 
\begin{figure}[ht]
\centering
\includegraphics[width=7 cm]{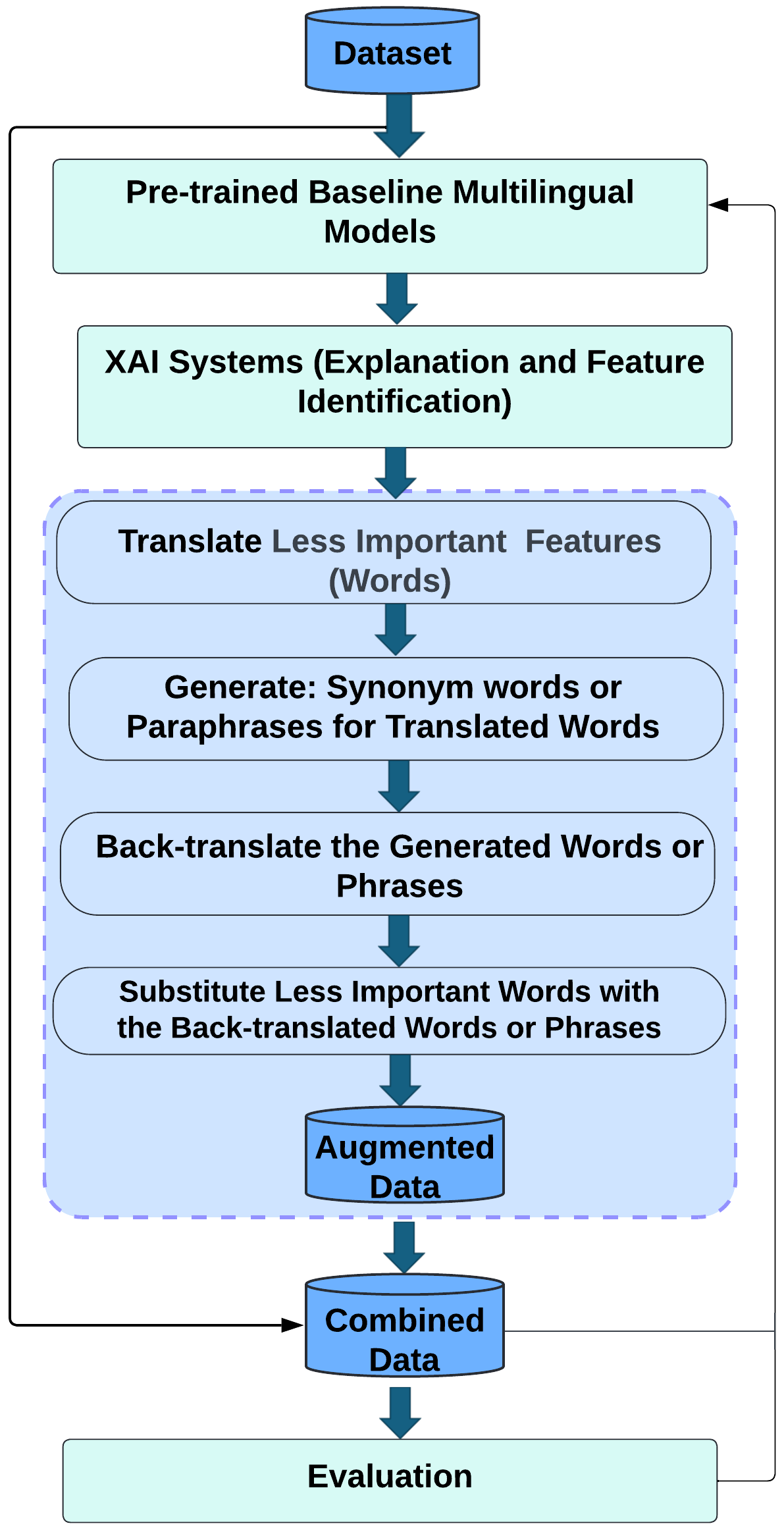}
\caption{XAI-guided and controlled context-aware data augmentation framework: An iterative XAI-guided feature identification data augmentation approach.}
\label{XAI_guided_model}
\end{figure}

Figure \ref{XAI_guided_model} provides a high-level overview of our XAI-guided context-aware data augmentation methodology. The augmentation process starts by leveraging pre-trained multilingual transformer-based encoder models as the baseline model. To better understand the model’s decision-making, XAI techniques are integrated to identify and highlight the less important features in the dataset. Once these less influential words are identified, the data augmentation process begins, involving multiple steps.

In the XAI Synonym Replacement with Back Translation (XAI-SR-BT) approach, the less influential words identified by the XAI method are first translated into a high-resource language (English) using the Google Translate API.  For each translated word, contextually appropriate synonyms are retrieved using the WordNet lexical database via the NLTK interface \cite{fellbaum1998wordnet, bird2009natural}. These synonyms are then back-translated into the original language using the same translation API. Finally, the original less important words in the input text are replaced with the back-translated synonyms. This method introduces controlled lexical variation while preserving semantic equivalence.

In the XAI Paraphrasing Replacement with Back Translation (XAI-PR-BT) approach, the translation process follows the same steps as in XAI-SR-BT. The fine-tuned version of the pre-trained PEGASUS model (tuner007/pegasus\_paraphrase) is then applied to generate paraphrases for the translated less influential words \cite{zhang2020pegasus}. However, other pre-trained models, such as GPT, can also be used for phrase generation. PEGASUS is better suited for this task because of its architecture. Unlike GPT, which is designed for next token prediction and open-ended generation, PEGASUS is pre-trained with a gap-sentence generation objective, making it more effective in producing semantically consistent and contextually relevant phrases \cite{zhang2020pegasus}. These paraphrased outputs are back-translated into the original language using the same translation API. Finally, the original less important words in the input text are replaced with the back-translated paraphrases. This technique enables more fluent and context-aware augmentation, thereby improving the model's robustness and generalization across linguistic variations.

The effectiveness of these augmentation strategies is evaluated by assessing the models’ performance metrics and the explanations provided by the XAI techniques.
Our approach incorporates an iterative feedback refinement loop into the XAI-guided augmentation pipeline, as illustrated in Figure \ref{XAI_guided_model}, to ensure the effectiveness and quality of augmentation.  After performing the augmentation process, which is carried out through XAI-SR-BT  and XAI-PR-BT, we dynamically verify whether the replacements for the selected less important words were successfully applied. If replacements are missing due to the unavailability of appropriate synonyms or paraphrases, the process loops back and dynamically adjusts the selection threshold for less important words identified by the XAI method (e.g., increasing the threshold from 20\% to 30\%). 

Secondly, once the augmented data is generated, it is combined with the original dataset and used to retrain the model. The model is then evaluated using the accuracy and the F1 score metrics, along with explanation insights to assess the quality and interpretability of the predictions. If the evaluation results indicate insufficient improvement, the process enters a refinement phase again. In this iterative loop stage, we adjust the translation API, the paraphrasing model, or the XAI technique. This iterative process continues until the model demonstrates meaningful improvements. This flexibility is a key benefit of our methodology. 

\begin{table*}[ht]
\centering

\begin{tabular}{lcccc}
\toprule
\textbf{Languages} & \textbf{Size} & \textbf{Average length} & \textbf{Data source} \\
\midrule
\multicolumn{4}{c}{\textbf{Hate Speech Dataset}} \\ \midrule
Amharic  & 14,643 & 17.5 & \cite{ayele2023exploring}  \\
Arabic  & 3,353 & 13.6 & \cite{ousidhoum2019multilingual}  \\
Hindi & 8,192 & 25.6 & \cite{bhardwaj2020hostility} \\
Italian & 8,100 & 24.7 & \cite{bosco2018overview} \\
Portuguese &  3,059 & 15.7 & \cite{fortuna2019hierarchically} \\
Spanish & 6,600 & 21.3 & \cite{basile2019semeval}  \\
\midrule
\multicolumn{4}{c}{\textbf{Sentiment Analysis Dataset}} \\ \midrule
Amharic (Ethiopia)  & 9,480  & 19.7 & \cite{muhammad-etal-2023-afrisenti}  \\
Hausa (Nigeria)  & 22,153  & 18  & \cite{muhammad-etal-2023-afrisenti}  \\
Kinyarwanda (Rwanda) & 5,155 & 14.6 & \cite{muhammad-etal-2023-afrisenti} \\
Swahili (Kenya) & 3,011 & 14 & \cite{muhammad-etal-2023-afrisenti}\\
\bottomrule
\end{tabular}
\caption{An overview of hate speech and sentiment analysis dataset statistics for multiple languages used in this study, including dataset size, average sentence length, and data sources.}
\label{table:data}
\end{table*}

\subsection{Datasets }
We utilized six hate speech datasets spanning multiple languages, ranging from high-resource Spanish to low-resource Amharic, including Arabic, Hindi, Italian, and Portuguese. Additionally, to demonstrate the reproducibility of our approach, we employed four sentiment analysis datasets in Amharic, Hausa, Kinyarwanda, and Swahili.  These datasets are carefully selected to evaluate the effectiveness of XAI-guided context-aware data augmentation techniques in high, medium, and low-resource languages. Each dataset differs in size, linguistic structure, and complexity, providing a robust and comprehensive benchmark for assessing the impact of our approach on model performance. This diversity enables us to investigate how well our proposed method generalizes across languages and varying levels of resource availability. Table \ref{table:data} presents an overview of the characteristics and statistics of each dataset used in our experiments. 

\subsection{Models}
Our model selection strategy is driven by two primary objectives: optimizing the data augmentation process and thoroughly evaluating the impact of the augmented data on model performance. To guide the augmentation process, we selected XLM-Roberta (XLM-R) \cite{conneau2019unsupervised} as our baseline model due to its proven robustness and exceptional performance in cross-lingual tasks. We also considered other pre-trained multilingual models as they provide a strong foundation for language understanding and can significantly enhance performance in various language scenarios. We fine-tune these models on both the original and augmented datasets to adapt them to specific downstream tasks. This allows us to evaluate the impact of data augmentation on task-specific performance.

To evaluate the impact of our augmented data, we utilized XLM-R and mBERT \cite{pires2019multilingual}, allowing for a comprehensive assessment of the effectiveness of our augmentation strategies across different architectures and languages. This dual-model evaluation enabled us to systematically analyze improvements in generalization, accuracy, and robustness across diverse languages, offering more profound insights into how the augmented data influences model performance in different language scenarios. By employing a multi-model evaluation approach, we validate our augmentation’s contribution to enhancing multilingual model performance and better understand the benefits of XAI-guided methodologies in different language settings.

\subsection{Selection of Explainability Techniques}
The selection of the appropriate XAI method is crucial and must be based on factors such as the interpretability needs, model architecture, and input data type \cite{yigezu2024ethio}. Several categories of XAI techniques provide distinct ways to interpret complex models, including perturbation-based methods like SHAP \cite{ribeiro2016should}, gradient-based approaches such as Integrated Gradients (IG) \cite{sundararajan2017axiomatic}, Layer-wise Relevance Propagation \cite{bach2015pixel}, and attention mechanism visualizations \cite{honnibal2017spacy}.

IG provides more accurate and interpretable word-level feature attributions than perturbation-based approaches, such as SHAP and LIME \cite{zhuo2024ig}. IG is ideal for text-based models because it captures the contextual influence of each word on the model's predictions \cite{ansari2024data}. IG is well-suited for handling long sequential text inputs, as it leverages the full context of the input to compute attribution scores without the scalability limitations associated with perturbation-based methods such as SHAP and LIME \cite{sundararajan2017axiomatic}. Unlike SHAP and LIME, which rely on repeated input perturbations and extensive sampling to estimate feature importance \cite{bhattacharya2022applied}, IG performs a single pass using gradient computations along a defined path from baseline to the actual input \cite{sundararajan2017axiomatic}. This makes IG significantly more computationally efficient and scalable for large-scale NLP tasks involving transformer-based models and long-form text data \cite{atanasova2024diagnostic}.

 IG ensures that the selected features for augmentation have minimal impact on the model’s overall decision-making process. IG is less sensitive to noise than perturbation-based methods like SHAP and LIME \cite{sundararajan2017axiomatic}, which can produce inconsistent feature importance scores due to their reliance on random perturbations. Perturbation-based XAI methods are also computationally expensive and often struggle with longer text sequences, leading to scalability issues and impractical for large-scale text data \cite{holzinger2022explainable, lundberg2017unified}.
Attention-based visualization techniques highlight words based on the attention weights assigned to each token by the model. However, these weights do not necessarily correlate with the true importance of features for the model’s final prediction, which makes them unreliable for guiding XAI-guided context-aware data augmentation \cite{serrano2019attention}. For instance, sentiment-bearing words like “excellent” and “poor” are crucial for determining classification in a classification task. In contrast, structural words like “had” or “but” might receive higher attention scores due to their contextual role, even though they contribute minimally to the text classification. Relying solely on attention scores for feature selection could lead to the preservation of irrelevant words and result in suboptimal data augmentation strategies.

We further contextualize our selection of IG by referencing our prior study (Mersha et al., 2025), in which we systematically evaluated the effectiveness of various XAI techniques across encoder-based language models. Our findings demonstrated that gradient-based techniques outperform SHAP, LIME, and attention visualization methods in terms of scalability and robustness in both short- and long-text inputs, particularly as the size of the model parameter increases \cite{mersha2025evaluating, mersha2025unified}.
In contrast, Integrated Gradients accurately quantify the contribution of each word, distinguishing between critical and non-essential features \cite{sundararajan2017axiomatic}. This capability allows us to modify or replace less important words while preserving the core meaning of the text \cite{kolesnikovadetecting}. Thus, we determine that IG is the optimal choice for XAI-guided context-aware data augmentation due to its interpretability, computational efficiency, and ability to maintain the integrity of the augmented data \cite{ismail2021improving}.

\subsection{XAI-Guided Data Augmentation Strategy}
The augmentation process begins using pre-trained multilingual models as the baseline model. We apply feature-based XAI techniques, such as Integrated Gradients, to explain the model’s decision-making process and identify the features in the data that have minimal impact on its predictions. Building on this, we proposed two XAI-guided context-aware data augmentation methods: XAI Synonym Replacement with Back Translation and Paraphrasing Replacement with Back Translation  These methods leverage explainability to selectively augment the data by targeting features that contribute the least to the model's predictions \cite{yigezuodio}. In both approaches, we employed the Integrated Gradients explainability technique to compute the importance score \(I(f_i)\) for each feature (word) \(f_i\) within our dataset \(D_{\text{original}}\). Integrated Gradients measure each feature's contribution to the model's prediction. To select the least important top-\(k\) features from a total of \(K\) features in the input example, we assumed that each feature has an importance score \(I(f_i)\), where \(f_i\) is the \(i\)-th feature, and \(I(f_i)\) represents its importance. The features were ranked in ascending order based on their importance scores, resulting in an ordered list \(f_{(1)}, f_{(2)}, \dots, f_{(K)}\), where \(I(f_{(1)}) \leq I(f_{(2)}) \leq \dots \leq I(f_{(K)})\). From this list, we selected the top-\(k\) least important features for targeted data augmentation, denoted as \(S_k = \{ f_{(1)}, f_{(2)}, \dots, f_{(k)} \}\). The value of $k$ is determined by input length, with shorter texts having smaller $k$ values and longer texts having larger $k$ values.

In the XAI Synonym with Back Translation augmentation approach, we replaced synonyms through translation after identifying the least important words. For each word \(f_i \in S_k\), we translated it from the source language to English, represented as \(f_{\text{trans}} = \text{Translate}(f_i)\). We then replaced the translated word with its synonym, denoted as \(f_{\text{syn}} = \text{Synonym}(f_{\text{trans}})\), and back-translated the synonym back to source language as \(f_{\text{bt}} = \text{BackTranslate}(f_{\text{syn}})\). The original least important word \(f_i\) in each sample \(x \in D_{\text{original}}\) was replaced with its back-translated synonym \(f_{\text{bt}}\).

We then generated augmented samples by replacing the least important words \(S_k\) in each original sample \(x\) with the back-translated synonyms, represented as \(x_{\text{augmented}} = \text{Augment}(x, S_k)\). The augmented samples were collected into an augmented dataset, \(D_{\text{augmented}}\).

In the final step, we combined the augmented dataset \(D_{\text{augmented}}\) with the original dataset \(D_{\text{original}}\) to create the final dataset for further training and evaluation. This combined dataset is represented as \(D_{\text{combined}} = D_{\text{original}} \cup D_{\text{augmented}}\).

The XAI Paraphrasing with Back Translation method follows a similar process to the synonym replacement approach, but uses paraphrasing instead. After identifying the top-\( k \) least important features \( S_k \) using Integrated Gradients, each word \( f_i \in S_k \) was first translated from the source language into English (\( f_{\text{trans}} = \text{Translate}(f_i) \)). Instead of finding a synonym, we generated a paraphrase using pre-trained models, such as GPT or PEGASUS (\( f_{\text{para}} = \text{Paraphrase}(f_{\text{trans}}) \)). The paraphrased word was then back-translated into the original source language, resulting in \( f_{\text{bt}} = \text{BackTranslate}(f_{\text{para}}) \). Each original word \( f_i \) was replaced with its back-translated paraphrased version \( f_{\text{bt}} \), creating an augmented sample \( x_{\text{augmented}} = \text{Augment}(x, S_k) \) for each original sample \( x \in D_{\text{original}} \). As with the synonym replacement method, the augmented dataset \( D_{\text{augmented}} \) can also combined with the original dataset to form the final dataset \( D_{\text{combined}} = D_{\text{original}} \cup D_{\text{augmented}} \).

Through these approaches, we effectively generated augmented data that leverages XAI techniques to target the least important features (words). While any features can also be targeted, our approach focuses on the least important words, preserving essential ones to generate model-oriented augmented data. This ensures the core information is retained while enhancing diversity and improving model robustness and performance. The model learns to be less sensitive to noise and irrelevant inputs by altering less critical features. 

\begin{figure}[h!]
\centering
\includegraphics[width=8 cm]{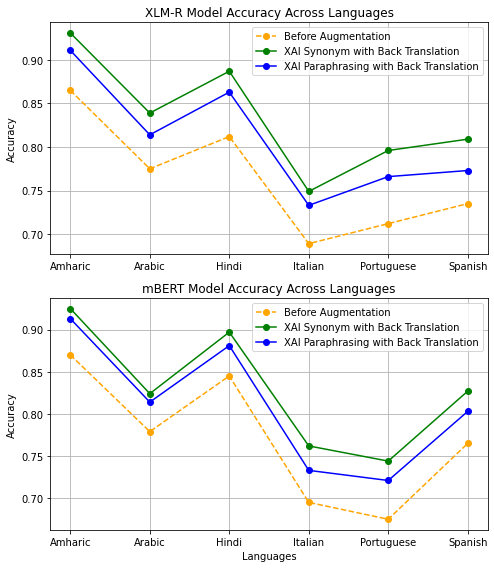}
\caption{ Accuracy comparison of XLM-R and mBERT models across multiple languages on a hate speech dataset, before and after XAI-guided context-aware data augmentation, showcasing the impact of augmented data on model performance.}
\label{fig:Accuracy}
\end{figure}

\begin{table*}[h!]
\centering
\scriptsize

     \begin{tabular}{p{1cm}p{1cm}p{0.8cm}p{0.8cm}p{0.8cm}p{0.8cm}p{0.8cm}p{0.8cm}p{0.8cm}p{0.8cm}p{0.8cm}p{0.8cm}}
    \toprule
    \multicolumn{2}{c}{} & \multicolumn{2}{c}{Before Augmentation} & \multicolumn{4}{c}{XAI Synonym with Back Translation} & \multicolumn{4}{c}{XAI Paraphrasing with Back Translation} \\ 
    \cmidrule(lr){3-4} \cmidrule(lr){5-8} \cmidrule(lr){9-12}
    Model & Datasets & Acc & F1 & Acc & F1 & $\Delta$Acc & $\Delta$F1 & Acc & F1 & $\Delta$Acc & $\Delta$F1 \\
    \midrule
    \multirow{6}{*}{\textbf{XLM-R}} 
    & Amharic     & 0.865 & 0.864 & 0.931 & 0.931 & +0.066 & +0.067 & 0.911 & 0.911 & +0.046 & +0.047 \\
    & Arabic      & 0.775 & 0.769 & 0.839 & 0.836 & +0.064 & +0.067 & 0.814 & 0.811 & +0.039 & +0.042 \\
    & Hindi       & 0.812 & 0.810 & 0.887 & 0.885 & +0.075 & +0.075 & 0.863 & 0.860 & +0.051 & +0.050 \\
    & Italian     & 0.689 & 0.678 & 0.749 & 0.748 & +0.060 & +0.070 & 0.733 & 0.733 & +0.044 & +0.055 \\
    & Portuguese  & 0.712 & 0.709 & 0.796 & 0.795 & +0.084 & +0.086 & 0.766 & 0.761 & +0.054 & +0.052 \\
    & Spanish     & 0.735 & 0.735 & 0.809 & 0.809 & +0.074 & +0.074 & 0.773 & 0.771 & +0.038 & +0.036 \\
    \midrule
    \multirow{6}{*}{\textbf{mBERT}} 
    & Amharic     & 0.870 & 0.872 & 0.925 & 0.925 & +0.055 & +0.053 & 0.913 & 0.912 & +0.043 & +0.040 \\
    & Arabic      & 0.779 & 0.766 & 0.824 & 0.823 & +0.045 & +0.057 & 0.814 & 0.812 & +0.035 & +0.046 \\
    & Hindi       & 0.845 & 0.834 & 0.897 & 0.892 & +0.052 & +0.058 & 0.881 & 0.880 & +0.036 & +0.046 \\
    & Italian     & 0.695 & 0.695 & 0.762 & 0.762 & +0.067 & +0.067 & 0.733 & 0.733 & +0.038 & +0.038 \\
    & Portuguese  & 0.675 & 0.669 & 0.744 & 0.742 & +0.069 & +0.073 & 0.721 & 0.711 & +0.046 & +0.042 \\
    & Spanish     & 0.765 & 0.765 & 0.827 & 0.822 & +0.062 & +0.057 & 0.803 & 0.795 & +0.038 & +0.030 \\
    \bottomrule
    \end{tabular}
    \caption{Impact of XAI-guided Synonym Replacement and Paraphrasing with Back Translation on XLM-R and mBERT models across different languages for the \textbf{hate speech detection task}. This table shows the improvements in Accuracy (Acc) and F1 score (F1) before and after augmentation, along with the corresponding delta values ($\Delta$Acc and $\Delta$F1). }
\label{table2:models_results}
\end{table*}

\begin{table*}[h!]
\centering
\scriptsize
\begin{tabular}{p{1cm}p{1cm}p{0.8cm}p{0.8cm}p{0.8cm}p{0.8cm}p{0.8cm}p{0.8cm}p{0.8cm}p{0.8cm}p{0.8cm}p{0.8cm}}
\toprule
\multicolumn{2}{c}{} & \multicolumn{2}{c}{Before Augmentation} & \multicolumn{4}{c}{XAI Synonym with Back Translation} & \multicolumn{4}{c}{XAI Paraphrasing with Back Translation} \\ 
\cmidrule(lr){3-4} \cmidrule(lr){5-8} \cmidrule(lr){9-12}
Model & Dataset & Acc & F1 & Acc & F1 & $\Delta$Acc & $\Delta$F1 & Acc & F1 & $\Delta$Acc & $\Delta$F1 \\
\midrule
\multirow{4}{*}{\textbf{XLM-R}} 
& Amharic     & 0.803 & 0.801 & 0.884 & 0.883 & 0.081 & 0.082 & 0.863 & 0.854 & 0.060 & 0.053 \\
& Hausa       & 0.827 & 0.825 & 0.907 & 0.896 & 0.080 & 0.071 & 0.875 & 0.874 & 0.048 & 0.049 \\
& Kinyarwanda & 0.706 & 0.700 & 0.764 & 0.761 & 0.058 & 0.061 & 0.749 & 0.741 & 0.043 & 0.041 \\
& Swahili     & 0.648 & 0.626 & 0.687 & 0.679 & 0.039 & 0.053 & 0.671 & 0.663 & 0.023 & 0.037 \\
\midrule
\multirow{4}{*}{\textbf{mBERT}} 
& Amharic     & 0.814 & 0.811 & 0.892 & 0.891 & 0.078 & 0.080 & 0.872 & 0.870 & 0.058 & 0.059 \\
& Hausa       & 0.819 & 0.814 & 0.914 & 0.913 & 0.095 & 0.099 & 0.884 & 0.881 & 0.065 & 0.067 \\
& Kinyarwanda & 0.717 & 0.710 & 0.776 & 0.774 & 0.059 & 0.064 & 0.751 & 0.743 & 0.034 & 0.033 \\
& Swahili     & 0.661 & 0.653 & 0.698 & 0.693 & 0.037 & 0.040 & 0.689 & 0.686 & 0.028 & 0.033 \\
\bottomrule
\end{tabular}
\caption{Impact of XAI-guided Synonym Replacement and Paraphrasing with Back Translation on XLM-R and mBERT models across four low-resource languages for the \textbf{sentiment analysis task}. The table shows the Accuracy (Acc) and F1 score (F1) before and after augmentation, along with the corresponding delta values ($\Delta$Acc and $\Delta$F1).}
\label{table3:models_results}
\end{table*}

\section{Experimental Results and Discussion}  
We conducted two experiments to generate augmented data using XAI-guided context-aware data augmentation techniques with the XLM-R model across six languages. The first experiment applied XAI-guided Synonym Replacement with Back Translation, while the second employed XAI-guided Paraphrasing with Back Translation.

We measured improvements in accuracy and F1 score using both XLM-R and mBERT models to evaluate the effectiveness of the XAI-guided augmentation techniques. The results from both experiments, presented in Table \ref{table2:models_results} and \ref{table3:models_results}, demonstrate that models trained on XAI-guided augmented data consistently outperformed models trained on non-augmented data. The XAI Synonym Replacement with Back Translation approach achieved the highest performance gains across all languages for both XLM-R and mBERT models, showing larger improvements in accuracy and F1 score compared to the XAI Paraphrasing with Back Translation technique. The XAI Synonym Replacement with the Back Translation approach captures semantic variations more effectively, leading to better generalization and stronger model performance. Although XAI Paraphrasing with Back Translation also produced positive results, its impact was generally smaller than XAI Synonym Replacement with the Back Translation approach, which may not add sufficient diversity to the augmented data, as shown in Figure \ref{fig:Accuracy}. 
 
 These performance gains can be attributed to XAI-guided context-aware data augmentations' more informed and targeted nature.
Unlike conventional methods that apply augmentations without accounting for feature (word) relevance for the decision-making process, which can introduce noise and distort the augmented data, our XAI-guided techniques leverage explainability methods, such as integrated gradients, to focus on altering less significant words in the input text. This selective approach ensures that the most critical features (words) are preserved while introducing controlled variability that enhances the model’s generalization ability.

\begin{table*}[h!]
\centering
\scriptsize

    \begin{tabular}{p{0.7cm}p{0.8cm}p{0.5cm}p{0.5cm}p{0.5cm}p{0.5cm}p{0.5cm}p{0.5cm}p{0.5cm}p{0.5cm}p{0.5cm}p{0.5cm}p{0.5cm}p{0.5cm}p{0.5cm}p{0.5cm}}
    \toprule
    \multicolumn{2}{c}{} & \multicolumn{2}{p{1.2cm}}{Before Augmentation} & \multicolumn{2}{p{1.3cm}}{Synonym Replacement} & \multicolumn{2}{p{1.3cm}}{Back Translation} & \multicolumn{2}{p{1.4cm}}{Contextual Augmentation} & \multicolumn{2}{p{1.3cm}}{Adversarial Examples} & \multicolumn{2}{p{1cm}}{\textbf{XAI-SR-BT }(ours)} & \multicolumn{2}{p{1cm}}{\textbf{XAI-PR-BT} (ours)} \\ 
    \cmidrule(lr){3-4} \cmidrule(lr){5-6} \cmidrule(lr){7-8} \cmidrule(lr){9-10} \cmidrule(lr){11-12} \cmidrule(lr){13-14} \cmidrule(lr){15-16}
    Model & Datasets & Acc & F1 & Acc & F1 & Acc & F1 & Acc & F1 & Acc & F1 & Acc & F1 & Acc & F1 \\
    \midrule
    \textbf{XLMR} & Amharic & 0.865 & 0.864 & 0.872 & 0.871 & \underline {0.883} & 0.882 & 0.869 & 0.868 & 0.870 & 0.870 & \textbf{0.931} & 0.931 & \textbf{\textit{0.911}} & 0.911 \\
    & Arabic & 0.775 & 0.769 & 0.790 & 0.788 & \underline {0.807} & 0.804 & 0.778 & 0.776 & 0.782 & 0.781 & \textbf{0.839} & 0.836 & \textbf{\textit{0.814}} & 0.811 \\
    & Hindi & 0.812 & 0.810 & 0.828 & 0.827 & \underline {0.843} & 0.843 & 0.824 & 0.823 & 0.819 & 0.819 & \textbf{0.887} & 0.885 & \textbf{\textit{0.863}} & 0.860 \\
    & Italian & 0.689 & 0.678 & 0.697 & 0.693 & \underline {0.716} & 0.713 & 0.692 & 0.691 & 0.701 & 0.700 & \textbf{0.749} & 0.748 & \textbf{\textit{0.733}} & 0.733 \\
    & Portuguese & 0.712 & 0.709 & 0.726 & 0.724 & \underline {0.747} & 0.742 & 0.718 & 0.718 & 0.720 & 0.719 & \textbf{0.796} & 0.795 & \textbf{\textit{0.766}} & 0.761 \\
    & Spanish & 0.735 & 0.735 & 0.747 & 0.747 & \underline {0.756} & 0.756 & 0.741 & 0.740 & 0.742 & 0.739 & \textbf{0.809} & 0.809 & \textbf{\textit{0.773}} & 0.771 \\
    
    \midrule
    \textbf{mBERT} & Amharic & 0.870 & 0.872 & 0.887 & 0.880 & \underline {0.894} & 0.890 & 0.877 & 0.870 & 0.868 & 0.861 & \textbf{0.925} & 0.925 & \textbf{\textit{0.913}} & 0.912 \\
    & Arabic & 0.779 & 0.766 & 0.792 & 0.789 & \underline {0.798} & 0.797 & 0.789 & 0.776 & 0.782 & 0.778 & \textbf{0.824} & 0.823 & \textbf{\textit{0.814}} & 0.812 \\
    & Hindi & 0.845 & 0.834 & 0.867 & 0.859 & \underline {0.871} & 0.871 & 0.855 & 0.851 & 0.850 & 0.847 & \textbf{0.897} & 0.892 & \textbf{\textit{0.881}} & 0.880 \\
    & Italian & 0.695 & 0.695 & 0.708 & 0.708 & \underline {0.724} & 0.719 & 0.699 & 0.696 & 0.700 & 0.700 & \textbf{0.762} & 0.762 & \textbf{\textit{0.733}} & 0.733 \\
    & Portuguese & 0.675 & 0.669 & 0.686 & 0.683 & \underline {0.694} & 0.694 & 0.679 & 0.672 & 0.680 & 0.677 & \textbf{0.744} & 0.742 & \textbf{\textit{0.721}} & 0.711 \\
    & Spanish & 0.765 & 0.765 & 0.774 & 0.774 & \underline {0.786} & 0.780 & 0.768 & 0.758 & 0.770 & 0.761 & \textbf{0.827} & 0.822 & \textbf{\textit{0.803}} & 0.795 \\ 
        
    \bottomrule
    \end{tabular}
    \caption{Performance comparison of conventional and XAI-guided context-aware data augmentation techniques on XLM-R and mBERT models across different languages for the \textbf{hate speech detection task}. The table presents the accuracy (Acc) and F1 score (F1), showing changes before and after applying different augmentation techniques. \textit{ \small  Note: XAI-SR-BT refers to XAI Synonym Replacement with Back Translation, and XAI-PR-BT refers to Paraphrasing Replacement with Back Translation }}
\label{table4:comparison}
\end{table*}

\begin{table*}[h!]
\centering
\scriptsize

    \begin{tabular}{p{0.7cm}p{1cm}p{0.5cm}p{0.5cm}p{0.5cm}p{0.5cm}p{0.5cm}p{0.5cm}p{0.5cm}p{0.5cm}p{0.5cm}p{0.5cm}p{0.5cm}p{0.5cm}p{0.5cm}p{0.5cm}}
    \toprule
    \multicolumn{2}{c}{} & \multicolumn{2}{p{1.2cm}}{Before Augmentation} & \multicolumn{2}{p{1.3cm}}{Synonym Replacement} & \multicolumn{2}{p{1.3cm}}{Back Translation} & \multicolumn{2}{p{1.4cm}}{Contextual Augmentation} & \multicolumn{2}{p{1.3cm}}{Adversarial Examples} & \multicolumn{2}{p{1cm}}{\textbf{XAI-SR-BT }(ours)} & \multicolumn{2}{p{1cm}}{\textbf{XAI-PR-BT} (ours)} \\ 
\cmidrule(lr){3-4} \cmidrule(lr){5-6} \cmidrule(lr){7-8} \cmidrule(lr){9-10} \cmidrule(lr){11-12} \cmidrule(lr){13-14} \cmidrule(lr){15-16}
Model & Dataset & Acc & F1 & Acc & F1 & Acc & F1 & Acc & F1 & Acc & F1 & Acc & F1 & Acc & F1 \\
\midrule
\multirow{4}{*}{\textbf{XLM-R}} 
& Amharic     & 0.803 & 0.801 & 0.834 & 0.830 & 0.843 & 0.842 & 0.819 & 0.814 & 0.828 & 0.811 & \textbf{0.884} & 0.883 & \textbf{\textit{0.863}} & 0.854 \\
& Hausa       & 0.827 & 0.825 & 0.847 & 0.844 & 0.851 & 0.850 & 0.837 & 0.836 & 0.844 & 0.844 & \textbf{0.907} & 0.896 & \textbf{\textit{0.875}} & 0.874 \\
& Kinyarwanda & 0.706 & 0.700 & 0.728 & 0.724 & 0.737 & 0.737 & 0.710 & 0.710 & 0.716 & 0.715 & \textbf{0.764} & 0.761 & \textbf{\textit{0.749}} & 0.741 \\
& Swahili     & 0.648 & 0.626 & 0.651 & 0.650 & 0.660 & 0.660 & 0.647 & 0.643 & 0.659 & 0.658 & \textbf{0.687} & 0.679 & \textbf{\textit{0.671}} & 0.663 \\
\midrule
\multirow{4}{*}{\textbf{mBERT}} 
& Amharic     & 0.814 & 0.811 & 0.847 & 0.844 & 0.855 & 0.855 & 0.828 & 0.822 & 0.833 & 0.832 & \textbf{0.892} & 0.891 & \textbf{\textit{0.872}} & 0.870 \\
& Hausa       & 0.819 & 0.814 & 0.854 & 0.851 & 0.861 & 0.860 & 0.836 & 0.834 & 0.840 & 0.840 & \textbf{0.914} & 0.913 & \textbf{\textit{0.884}} & 0.881 \\
& Kinyarwanda & 0.717 & 0.710 & 0.731 & 0.730 & 0.744 & 0.745 & 0.721 & 0.721 & 0.724 & 0.724 & \textbf{0.776} & 0.774 & \textbf{\textit{0.751}} & 0.743 \\
& Swahili     & 0.661 & 0.653 & 0.671 & 0.670 & 0.680 & 0.680 & 0.665 & 0.662 & 0.669 & 0.668 & \textbf{0.698} & 0.693 & \textbf{\textit{0.689}} & 0.686 \\
\bottomrule
\end{tabular}
\caption{Performance comparison of conventional and XAI-guided context-aware data augmentation techniques on XLM-R and mBERT models across four low-resource languages for the \textbf{sentiment analysis task}. The table presents the accuracy (Acc) and F1 score (F1), showing changes before and after applying different augmentation techniques. \textit{ \small  Note: XAI-SR-BT refers to XAI Synonym Replacement with Back Translation, and XAI-PR-BT refers to Paraphrasing Replacement with Back Translation }}
\label{table5:comparison}
\end{table*}

We ensured that critical features remained consistent across datasets, which helped us maintain semantic integrity and prevent distortions that could undermine the model's prediction. When discrepancies in feature importance were detected—such as shifts in the importance of crucial features—we adjusted the augmentation parameter, including the value of $k$ in the $top-k$ least important feature selection parameter, to ensure that the process accurately targeted and preserved relevant features. To ensure meaningful augmentation while preserving semantic integrity, we limited 
$k$ to 30\% of the total features or words in our experiment, selecting them from the least to the most important ranked words based on the input length of our dataset, as shown in Table~\ref{table:data}. The value of 
$k$ was dynamically adjusted for each input text based on its length, with shorter inputs assigned a smaller 
$k$ and longer inputs assigned a larger 
$k$. Additionally, we can generate diverse augmented datasets to optimize performance by varying the value of $k$ and evaluating iteratively. The augmentations are driven by the model’s internal understanding of the input. As a result, XAI-guided context-aware data augmentation techniques consistently improve accuracy and F1 scores across several languages, outperforming conventional augmentation methods that often yield limited improvements, as shown in Table \ref{table4:comparison} and \ref{table5:comparison}. 

The performance differences in XAI-guided data augmentation between low-resource and high-resource languages can be attributed to linguistic characteristics and differences in corpus size. Multilingual models may tend to be biased toward high-resource languages due to factors such as syntactic regularity, lexical richness, and larger corpus sizes. These factors enable high-resource languages to benefit more from augmentation techniques, resulting in greater improvements in model performance. As shown in Table \ref{table2:models_results} and \ref{table3:models_results}, low-resource languages show lower gains in both accuracy and F1 scores. For instance, Swahili and Kinyarwanda achieved only 2.8\% and 3.4\% accuracy improvements, respectively, on sentiment analysis tasks when using the XAI Paraphrasing with Back Translation (XAI-PR-BT) approach. In contrast, high-resource languages such as Hindi, supported by richer lexical resources and larger corpora, demonstrated significant improvements, for example, a 7.5\% gain in accuracy on the hate speech task using the XAI Synonym Replacement with Back Translation (XAI-SR-BT) method.

These performance differences are further influenced by linguistic characteristics such as morphological complexity, syntactic variability, and limited corpus availability. Morphologically rich languages, such as many low-resource Indigenous Australian languages studied in \cite{tosolini2025data, tosolini2025data}, often exhibit complex inflectional and derivational systems, which create challenges for data augmentation techniques designed for high-resource languages with simpler morphology. Syntactic diversity, such as the order of free words in some languages, can diminish the effectiveness of text-based augmentation methods that assume more rigid syntactic structures \cite{li2022data}. Additionally, corpus size exacerbates these issues, as low-resource languages lack sufficient data to train robust models, rendering augmentation less effective compared to high-resource languages, where abundant data ensures better generalization\cite{nzeyimana2024low, solyman2023optimizing}.

\subsection{Comparison of XAI-Guided and Conventional Data Augmentation}
Tables \ref{table4:comparison} and \ref{table5:comparison} compare XAI-guided context-aware data augmentation techniques with conventional methods in multiple languages using the XLM-R and mBERT models. In our study, we compare the effectiveness of our XAI-guided context-aware data augmentation approach with four conventional techniques, which serve as benchmarks to evaluate our proposed method. Back Translation involves translating the text into another language and then back into the original language to generate variations \cite{feldman2020neural}. Random Synonym Replacement with Back Translation substitutes words in the original text with synonyms using translation \cite{beddiar2021data}. Contextual Augmentation employs pre-trained language models, such as mBERT and XLM-R, to replace words based on their context \cite{kobayashi2018contextual}. Adversarial examples with back translation introduce small perturbations or noise into the text \cite{ebrahimi2017hotflip}.  
Across all datasets, as shown in Table \ref{table2:models_results} and \ref{table3:models_results}, both XAI-guided context-aware data augmentation methods—XAI Synonym Substitution with Back Translation and XAI Paraphrasing with Back Translation—consistently outperform the baseline results before augmentation and all other convention augmentation methods. XAI Synonym Substitution with Back Translation provides more significant gains in most cases, showing more substantial improvements in accuracy and F1 score. This highlights the effectiveness of integrating explainability techniques into data augmentation to generate meaningful and diverse text variations that improve model performance across multiple datasets. 

Our XAI-guided context-aware data augmentation techniques, such as XAI Synonym Replacement with Back Translation and XAI Paraphrasing with Back Translation, show a clear performance advantage over conventional augmentation methods, including Synonym Replacement, Back Translation, Contextual Augmentation, and Adversarial Examples, as shown in Table \ref{table4:comparison} and \ref{table5:comparison}.

Appendix \ref{sample Augmented data} illustrates examples of augmented data in Spanish and Amharic languages, along with their English translations, by using a synonym replacement with a back-translation approach.

\subsection{Evaluation}
We utilized two complementary evaluation approaches to determine the effectiveness of our proposed XAI-guided context-aware data augmentation method: model-based and XAI-based evaluation. The model-based approach involves assessing the overall model performance gains before and after augmentation. The XAI-based approach uses XAI techniques to evaluate changes in feature importance and interpretability before and after augmentation.

For the model-based evaluation, we employed XLM-R and mBERT as evaluation models to analyze the performance of both the original and the augmented datasets. By comparing the accuracy and F1-score metrics, we measured the impact of augmented data on model performance. This enabled us to quantify the gains achieved through augmentation regarding generalization and robustness across different language scenarios, as shown in Table \ref{table2:models_results} and \ref{table3:models_results}.

The XAI-based evaluation leveraged feature attribution methods to better understand model behavior on the original and augmented datasets. We applied the Integrated gradient XAI technique to determine whether critical features identified in the original data were preserved in the augmented versions. By comparing feature importance scores and XAI explanations before and after augmentation, we assessed the impact of our augmentation strategies on model interpretability.

Our approach is not static but rather an ongoing process of refinement. This iterative feedback loop of model performance evaluation and XAI-based insights allows us to refine the augmentation process continuously. By doing so, we optimize model accuracy and robustness and enhance interpretability, ensuring that our augmented data effectively supports the model prediction performance and explainability goals.

\section{Limitations and Future Direction} \vspace{-0.7em}
Despite the strong performance gains demonstrated by our XAI-guided context-aware data augmentation approach, some limitations remain. The method depends on the selected explainability technique. This may introduce computational overhead due to the integration of XAI methods into the baseline model. Moreover, the reliability of current XAI explanations can be inconsistent, and the quality of augmentation may vary depending on the chosen XAI method and translation tool. In future work, we plan to investigate more efficient and stable XAI techniques and extend our evaluation across a broader range of model architectures (language-specific models like AfriBERTa and AraBERT), languages, and tasks to analyze performance improvements. We also intend to explore the implications of our approach for multilingual fairness and adversarial robustness.

\section{Conclusion}  
Our study introduces a novel XAI-guided context-aware data augmentation approach that effectively addresses the limitations of conventional data augmentation methods, particularly by improving the performance of language models in different languages.  The proposed approach ensures that critical semantic information is preserved during augmentation by integrating XAI techniques with multilingual models to identify less model-influential features in the input text. This targeted strategy results in high-quality augmented data, significantly enhancing various language models' robustness, performance, and explainability, offering a promising future, particularly for low-resource language processing.
We demonstrate that XAI-guided context-aware data augmentation enhances the generalization of language models and provides transparency into the augmentation process. Through iterative refinement using feedback loops between XAI and applied models, the approach maintains contextual integrity, making it particularly suitable for addressing the diverse challenges of several languages.
Our experimental results confirm that the XAI-guided context-aware data augmentation approach consistently outperforms existing or conventional augmentation methods, achieving higher accuracy and F1 scores across multiple language datasets. By integrating explainability techniques and language models into the data augmentation, we bridge the gap between enhancing model performance and maintaining interpretability.
Our study sets the stage for the broader application of XAI in enhancing various language models in different languages.

\bibliographystyle{elsarticle-num}
\bibliography{bib.bib}

\appendix

\begin{figure*}[p]
\centering
\includegraphics[width=1\textwidth, height=0.95\textheight]{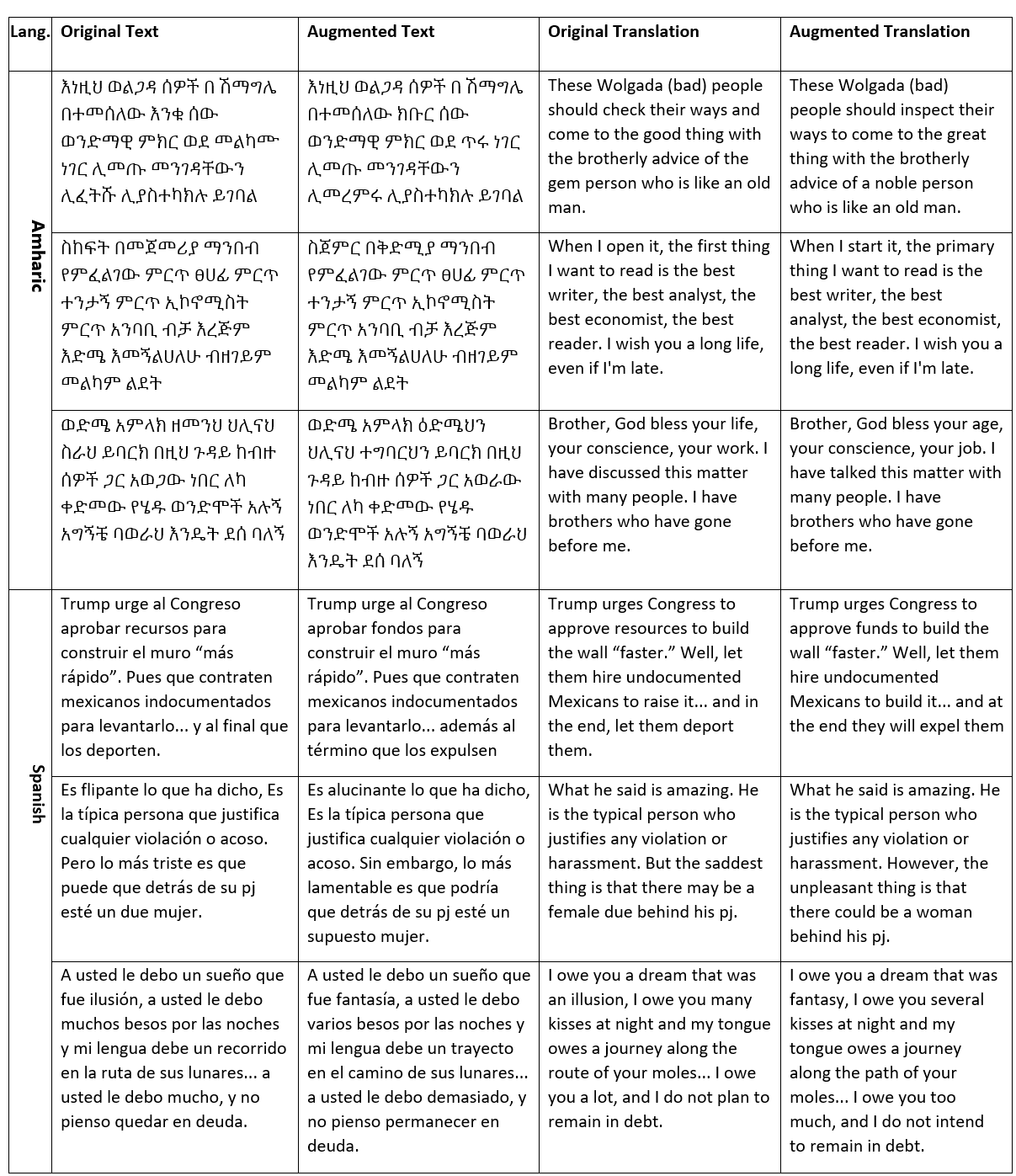}
\caption{Sample Spanish and Amharic augmented data with translations from the Hate Speech dataset using the Synonym Replacement with Back Translation Approach. \textcolor{red}{These samples include representative hate speech examples, which may contain offensive language, solely for research and analysis in the context of hate speech detection and data augmentation}.}
\label{sample Augmented data}
\end{figure*}

\section{Sample Augmented Data \ref{sample Augmented data}}

\end{document}